\documentclass[11pt,letterpaper]{article}
\usepackage{ijcnlp2017}
\usepackage{times}
\usepackage{latexsym}

\usepackage{listings}
\usepackage{multirow}
\usepackage{graphicx}
\graphicspath{ {images/} }
\usepackage{subcaption}
\usepackage{makecell}
\usepackage{amsmath}
\usepackage{footnote}
\makesavenoteenv{tabular}
\makesavenoteenv{table}
\usepackage{tablefootnote}

\ijcnlpfinalcopy

\title{What does Attention in Neural Machine Translation\\
 Pay Attention to?}

\author{Hamidreza Ghader \and Christof Monz\\
{ Informatics Institute, University of Amsterdam, The Netherlands}\\
  {\tt h.ghader, c.monz@uva.nl}}

\date{}

\begin{document}

\maketitle

\begin{abstract}
Attention in neural machine translation provides the possibility to encode relevant parts of the source sentence at each translation step. As a result, attention is considered to be an alignment model as well. However, there is no work that specifically studies attention and provides analysis of what is being learned by attention models. Thus, the question still remains that how attention is similar or different from the traditional alignment. In this paper, we provide detailed analysis of attention and compare it to traditional alignment. We answer the question of whether attention is only capable of modelling translational equivalent or it captures more information. We show that attention is different from alignment in some cases and is capturing useful information other than alignments.  
\end{abstract}

\section{Introduction}

Neural machine translation (NMT) has gained a lot of attention recently due to its substantial improvements in machine translation quality achieving state-of-the-art performance for several languages~\cite{luong-EtAl:2015:ACL-IJCNLP, jean-EtAl:2015:ACL-IJCNLP, wu2016google}. The core architecture of neural machine translation models is based on the general encoder-decoder approach~\cite{sutskever2014sequence}. Neural machine translation is an end-to-end approach that learns to encode source sentences into distributed representations and decode these representations into sentences in the target language.  Among the different neural MT models, attentional NMT~\cite{bahdanau-EtAl:2015:ICLR, DBLP_journals_corr_LuongPM15} has become popular due to its capability to use the most relevant parts of the source sentence at each translation step. This capability also makes the attentional model superior in translating longer sentences~\cite{bahdanau-EtAl:2015:ICLR, DBLP_journals_corr_LuongPM15}. 

\begin{figure}[thb]
\centering
\includegraphics[scale=0.28]{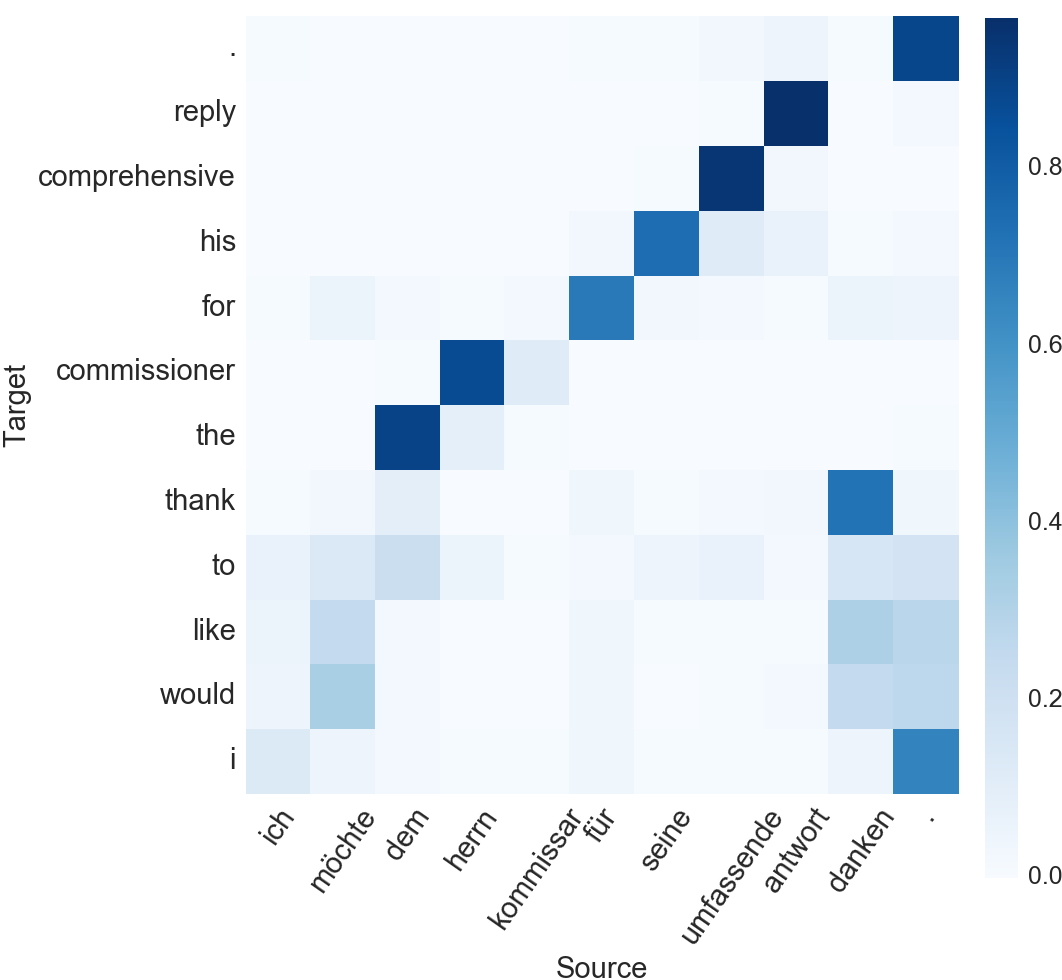}
\caption{Visualization of the attention paid to the relevant parts of the source sentence for each generated word of a translation example. See how the attention is `smeared out' over multiple source words in the case of ``would" and ``like".}
\label{fig:heatmap}
\end{figure}

Figure~\ref{fig:heatmap} shows an example of how attention uses the most relevant source words to generate a target word at each step of the translation. In this paper we focus on studying the relevance of the attended parts, especially cases where attention is `smeared out' over multiple source words where their relevance is not entirely obvious, see, e.g., ``would" and ``like" in Figure~\ref{fig:heatmap}. Here, we ask whether these are due to errors of the attention mechanism or are a desired behavior of the model.

Since the introduction of attention models in neural machine translation~\cite{bahdanau-EtAl:2015:ICLR} various modifications have been proposed~\cite{DBLP_journals_corr_LuongPM15, cohn-EtAl:2016:N16-1, liu-EtAl:2016:COLING}. However, to the best of our knowledge there is no study that provides an analysis of what kind of phenomena is being captured by attention. There are some works that have looked to attention as being similar to traditional word alignment~\cite{alkhouli-EtAl:2016:WMT, cohn-EtAl:2016:N16-1, liu-EtAl:2016:COLING, chen2016guided}. Some of these approaches also experimented with training the attention model using traditional alignments~\cite{alkhouli-EtAl:2016:WMT, liu-EtAl:2016:COLING, chen2016guided}.~\newcite{liu-EtAl:2016:COLING} have shown that attention could be seen as a reordering model as well as an alignment model.

In this paper, we focus on investigating the differences between attention and alignment and what is being captured by the attention mechanism in general. The questions that we are aiming to answer include: Is the attention model only capable of modelling alignment? And how similar is attention to alignment in different syntactic phenomena?
%Which syntactic phenomena cause the attention model to make errors? 
%
%In what situation the attention models attends distant places in the source side? Is this due to error of the attention model or is it a normal behaviour of the model to capture some syntactic characteristic? 
%

%\CM{this is repeated below: Our analysis shows that attention pattern depend on the type of the word that the model generates. For some word types, attention behaves more like alignments while for others it captures information going beyond word alignment. }
Our analysis shows that attention models traditional alignment in some cases more closely while it captures information beyond alignment in others. For instance, attention agrees with traditional alignments to a high degree in the case of nouns. However, it captures other information rather than only the translational equivalent in the case of verbs.

%To answer these questions, we train two different machine translation systems; one with an attention model without any input-feeding and the other with input-feeding \citep{DBLP_journals_corr_LuongPM15}. A detailed description of the models is given in the following sections. We investigate the results of translating using these systems and provide evidence for attention behaviour by comparing the statistics resulted from these systems against traditional alignments. 

%\CM{Rewrite in terms of contributions:
%In the following section we give a short discussion of the related works in the literature. Then, we briefly describe the architecture and the setup of our systems. In the next section, we talk about the experiments that we have done using multiple systems and give a comparison of their behaviour by providing statistics about translating different syntactic phenomena.}

%\CM{this is vague: 2) We compare two commonly used attention models and show that attention complying with alignments to a certain extent could be useful.}
%\CM{this is vague: 4) We provide evidence that a large amount of attention not paid to the alignment points of a word is distributed among other relevant words that possibly affect translation of the word. This particularly means that the difference of attention and alignment in those cases is a meaningful difference.}

This paper makes the following contributions: 1) We provide a detailed comparison of attention in NMT and word alignment. 2) We show that while different attention mechanisms can lead to different degrees of compliance with respect to word alignments, global compliance is not always helpful for word prediction.
%We also provide evidence that global compliance is not always helpful by showing that there are circumstances that closer compliance helps or hurts.
%
%show that closer compliance helps the translation quality, however, a full agreement is not helpful. 
3) We show that attention follows different patterns depending on the type of the word being generated. 4) We demonstrate that attention does not always comply with alignment. We provide evidence showing that the difference between attention and alignment is due to attention model capability to attend the context words influencing the current word translation.

\section{Related Work}
\label{sec:relatedWork}
%There are a dozen of works on attentional models in neural machine translation, although these models has first been used in neural machine translation quite recently \citep{bahdanau-EtAl:2015:ICLR}. However, to the best of our knowledge, there is no work that has studied the attention in the sense of its difference with the traditional alignment.

%\newcite{DBLP_journals_corr_LuongPM15} propose a global and a local attention model. They argue that their global attention model is similar to the original attention model while having simpler architecture and doing less computation. In their local model they predict an alignment point at each step and make the attention to focus to a window with the aligned point at the centre. They also propose an input-feeding attention model which has an input of its previous attentional decisions to maintain a coverage of the source side. 

\newcite{liu-EtAl:2016:COLING} investigate how training the attention model in a supervised manner can benefit machine translation quality. To this end they use traditional alignments obtained by running automatic alignment tools (GIZA++~\cite{och2003systematic} and fast\_align~\cite{dyer-chahuneau-smith:2013:NAACL-HLT}) on the training data and feed it as ground truth to the attention network. They report some improvements in translation quality arguing that the attention model has learned to better align source and target words. The approach of training attention using traditional alignments has also been proposed by others~\cite{chen2016guided, alkhouli-EtAl:2016:WMT}.~\newcite{chen2016guided} show that guided attention with traditional alignment helps in the domain of e-commerce data which includes lots of out of vocabulary (OOV) product names and placeholders, but not much in the other domains.~\newcite{alkhouli-EtAl:2016:WMT} have separated the alignment model and translation model, reasoning that this avoids propagation of errors from one model to the other as well as providing more flexibility in the model types and training of the models. They use a feed-forward neural network as their alignment model that learns to model jumps in the source side using HMM/IBM alignments obtained by using GIZA++. 

%Their approach also achieves mixed results motivating a more thorough analysis of attention models in NMT.

%\newcite{shi-padhi-knight:2016:EMNLP2016} do an analysis of what syntactic information is being learned by an end to end neural machine translation system. 
\newcite{shi-padhi-knight:2016:EMNLP2016} show that various kinds of syntactic information are being learned and encoded in the output hidden states of the encoder. The neural system for their experimental analysis is not an attentional model and they argue that attention does not have any impact for learning syntactic information.  However, performing the same analysis for morphological information, \newcite{belinkov2017neural} show that attention has also some effect on the information that the encoder of neural machine translation system encodes in its output hidden states. As part of their analysis they show that a neural machine translation system that has an attention model can learn the POS tags of the source side more efficiently than a system without attention.

Recently, \newcite{koehn2017six} carried out a brief analysis of how much attention and alignment match in different languages by measuring  the probability mass that attention gives to alignments obtained from an automatic alignment tool. They also report differences based on the most attended words.

The mixed results reported by \newcite{chen2016guided, alkhouli-EtAl:2016:WMT, liu-EtAl:2016:COLING} on optimizing attention with respect to alignments motivates a more thorough analysis of attention models in NMT.
%\HG{This combination of the previous works shows that it is important to have a deeper analysis of the differences and the similarities between attention and alignments.}
 
\section{Attention Models}
\label{sec:attentionModels}

This section provides a short background on attention and discusses two most popular attention models which are also used in this paper. The first model is a non-recurrent attention model which is equivalent to the ``global attention" method proposed by \newcite{DBLP_journals_corr_LuongPM15}. The second attention model that we use in our investigation is an input-feeding model similar to the attention model first proposed by \newcite{bahdanau-EtAl:2015:ICLR} and turned to a more general one and called \emph{input-feeding} by \newcite{DBLP_journals_corr_LuongPM15}. Below we describe the details of both models.

Both non-recurrent and input-feeding models compute a context vector $c_i$ at each time step. Subsequently, they concatenate the context vector to the hidden state of decoder and pass it through a non-linearity before it is fed into the softmax output layer of the translation network.

\begin{equation}
\tilde{h}_{t} = tanh(W_c[c_t;h^{\prime}_{t}])
\label{eq:attOutput}
\end{equation}

The difference of the two models lays in the way they compute the context vector. In the non-recurrent model, the hidden state of the decoder is compared to each hidden state of the encoder. Often, this comparison is realized as the dot product of vectors. Then the comparison result is fed to a softmax layer to compute the attention weight.
\begin{equation}
e_{t,i}= h^{T}_{i} h^{\prime}_{t}
\label{eq:hiddenStateComparison}
\end{equation}
\begin{equation}
\alpha_{t,i} = \frac{\mathrm{exp}(e_{t,i})}{\sum_{j=1}^{|x|} \mathrm{exp}(e_{t,j})}
\label{eq:attentionWeight}
\end{equation}

Here $h^{\prime}_{t}$ is the hidden state of the decoder at time $t$, $h_{i}$ is $i$th hidden state of the encoder and $|x|$ is the length of the source sentence. Then the computed alignment weights are used to compute a weighted sum over the encoder hidden states which results in the context vector mentioned above:

\begin{equation}
c_i = \sum_{i=1}^{|x|}\alpha_{t,i}h_i
\end{equation}

The input-feeding model changes the context vector computation in a way that at each step $t$ the context vector is aware of the previously computed context $c_{t-1}$. To this end, the input-feeding model feeds back its own $\tilde{h}_{t-1}$ to the network and uses the resulting hidden state instead of the context-independent $h^{\prime}_{t}$, to compare to the hidden states of the encoder. This is defined in the following equations:

\begin{equation}
h^{\prime\prime}_{t}=f(W[\tilde{h}_{t-1};y_{t-1}])
\label{eq:stackLSTM}
\end{equation}

\begin{equation}
e_{t,i} = h_{i}^T h^{\prime\prime}_{t} 
\label{eq:inputFeedingComp}
\end{equation} 

Here, $f$ is the function that the stacked LSTM applies to the input, $y_{t-1}$ is the last generated target word, and $\tilde{h}_{t-1}$ is the output of previous time step of the input-feeding network itself, meaning the output of Equation~\ref{eq:attOutput} in the case that context vector has been computed using $e_{t,i}$ from Equation~\ref{eq:inputFeedingComp}. 

\section{Comparing Attention with Alignment}
As mentioned above, it is a commonly held assumption that attention corresponds to word alignments. To verify this, we investigate whether higher consistency between attention and alignment leads to better translations.
 
 \begin{table}
\centering
\small
\begin{tabular}{|c|c|}
\hline
& RWTH data\\
\hline
\# of sentences & 508\\
\# of alignments & 10534 \\
\% of sure alignments & 91\%\\
\% of possible alignments & 9\%\\
\hline
\end{tabular}
\caption{\label{tbl:RWTHdataStatistics} Statistics of manual alignments provided by RWTH German-English data.}
\end{table}

\subsection{Measuring Attention-Alignment Accuracy}

In order to compare attentions of multiple systems as well as to measure the difference between attention and word alignment, we convert the hard word alignments into soft ones and use cross entropy between attention and soft alignment as a loss function. For this purpose, we use manual alignments provided by RWTH German-English dataset as the hard alignments. The statistics of the data are given in Table~\ref{tbl:RWTHdataStatistics}. We convert the hard alignments to soft alignments using Equation~\ref{cs:alignmentDef}. For unaligned words, we first assume that they have been aligned to all the words in the source side and then do the conversion.
%\CM{introduce RWTH dataset here}

%\CM{explain the soft alignment conversion here before the equation}

\begin{equation}
Al(x_i, y_t) =
\begin{cases}
\frac{1}{|A_{y_t}|} & \mathrm{if}\ x_i \in A_{y_t}\\
0 & \mathrm{otherwise}
\end{cases}
\label{cs:alignmentDef}
\end{equation}

Here $A_{y_t}$ is the set of source words aligned to target word $y_t$ and $|A_{y_t}|$ is the number of source words in the set.

After conversion of the hard alignments to soft ones, we compute the \emph{attention loss} as follows:

\begin{equation}
L_{At}(y_t) = -\sum_{i=1}^{|x|}{Al(x_i,y_t)\log(At(x_i,y_t))}
\label{eq:attnLoss}
\end{equation}

Here $x$ is the source sentence and $Al(x_i,y_t)$ is the weight of the alignment link between source word $x_i$ and the target word (see Equation~\ref{cs:alignmentDef}). $At(x_i,y_t)$ is the attention weight $\alpha_{t,i}$ (see Equation~\ref{eq:attentionWeight}) of the source word $x_i$, when generating the target word $y_t$ . 
%\CM{this is unclear: We do the conversion of the hard alignments to soft alignments by uniformly distributing the mass between all aligned words in the source side,} \HG{as shown in Equation~\ref{cs:alignmentDef}}.  

In our analysis, we also look into the relation between translation quality and the quality of the attention with respect to the alignments. For measuring the quality of attention, we use the attention loss defined in Equation~\ref{eq:attnLoss}. As a measure of translation quality, we choose the loss between the output of our NMT system and the reference translation at each translation step, which we call \emph{word prediction loss}. The word prediction loss for word $y_t$ is logarithm of the probability given in Equation~\ref{eq:NMToutput}. 

\begin{equation}
p_{nmt}(y_t\mid y_{<t}, x) = softmax(W_o\tilde{h}_{t})
\label{eq:NMToutput}
\end{equation}

Here $x$ is the source sentence, $y_t$ is target word at time step $t$, $y_{<t}$ is the target history given by the reference translation and $\tilde{h}_t$ is given by Equation~\ref{eq:attOutput} for either non-recurrent or input-feeding attention models.

\begin{figure}[thb]
\includegraphics[scale=0.73]{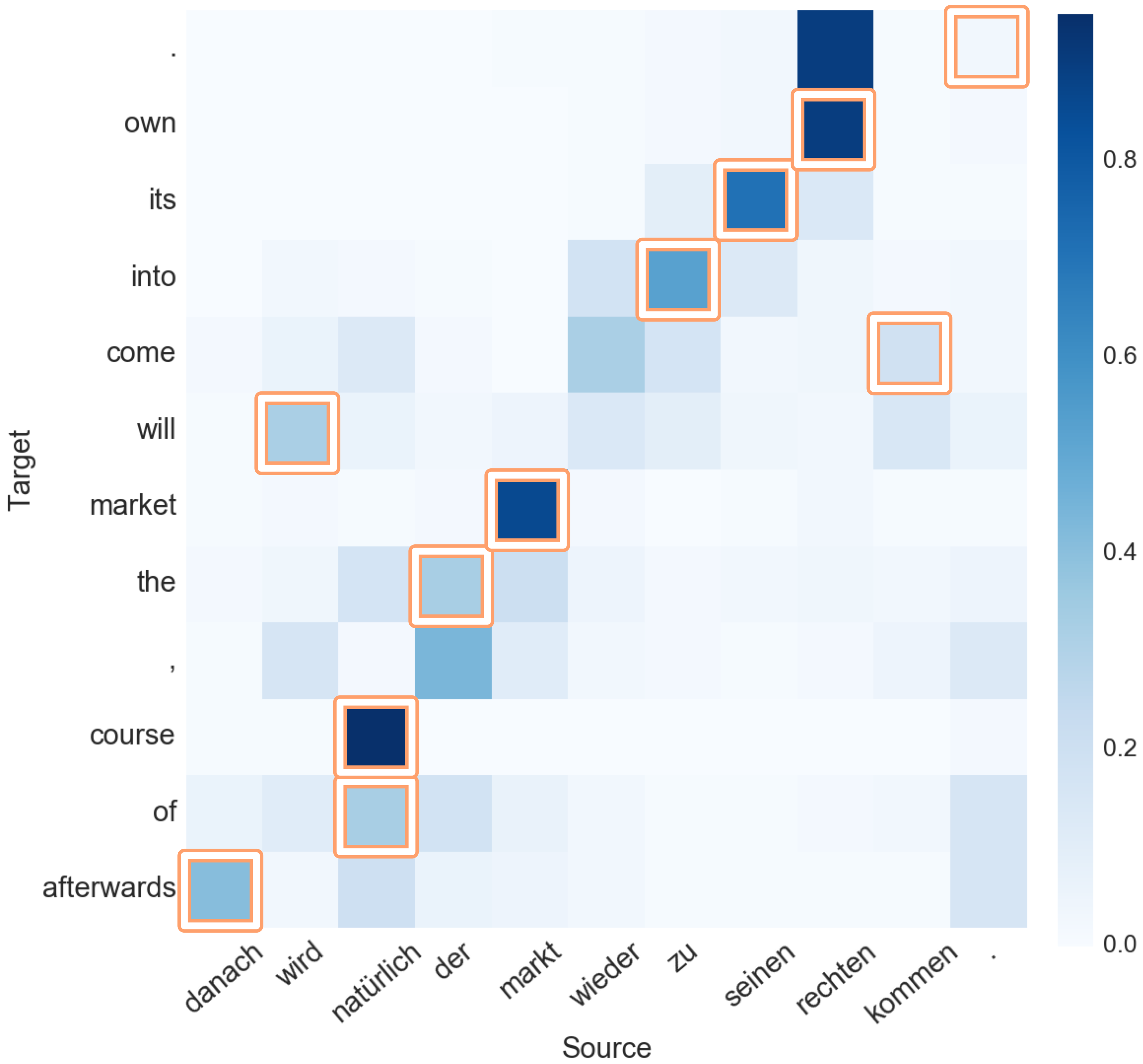}
\caption{An example of inconsistent attention and alignment. The outlined cells show the manual alignments from the RWTH dataset (see Table~\ref{tbl:RWTHdataStatistics}). See how attention is deviated from alignment points in the case of ``will" and ``come".}
\label{fig:attention_alignment}
\end{figure}

Spearman's rank correlation is used to compute the correlation between attention loss and word prediction loss:

\begin{equation}
\rho = \frac{\mathrm{Cov}(R_{L_{At}}, R_{L_{WP}})}{\sigma_{R_{L_{At}}}\sigma_{R_{L_{WP}}}}
\end{equation}
where $R_{L_{At}}$ and $R_{L_{WP}}$ are the ranks of the attention losses and word prediction losses, respectively, $\mathrm{Cov}$ is the covariance between two input variables, and $\sigma_{R_{L_{At}}}$ and $\sigma_{R_{L_{WP}}}$ are the standard deviations of $R_{L_{At}}$ and $R_{L_{WP}}$.

If there is a close relationship between word prediction quality and consistency of attention versus alignment, then there should be high correlation between word prediction loss and attention loss. Figure~\ref{fig:attention_alignment} shows an example with different levels of consistency between attention and word alignments. For the target words ``will" and ``come" the attention is not focused on the manually aligned word but distributed between the aligned word and other words. The focus of this paper is examining cases where attention does not follow alignment, answering the questions whether those cases represent errors or desirable behavior of the attention model.

\subsection{Measuring Attention Concentration}

As another informative variable in our analysis, we look into the attention concentration. While most word alignments only involve one or a few words, attention can be distributed more freely. We measure the concentration of attention by computing the entropy of the attention distribution:

\begin{equation}
E_{At}(y_t) = -\sum_{i=1}^{|x|}{At(x_i,y_t)\log(At(x_i,y_t))}
\label{eq:attnEntropy}
\end{equation}

%As in Equation~\ref{eq:attnLoss}, $W_s$ stands for the set of source words and $At(w_s)$ is the weight of the attention that the source word received when the system is generating the target word, $w_t$, \HG{as computed in equation~\ref{eq:attentionWeight} to be $\alpha_{t,s}$}.

\begin{table*}[t]
\centering
%\small
\begin{tabular}{|r|r|r|r|r|}
\hline
\multicolumn{1}{|c|}{System} & \multicolumn{1}{c|}{test2014} & \multicolumn{1}{c|}{test2015} & \multicolumn{1}{c|}{test2016}  & \multicolumn{1}{c|}{RWTH}\\
\hline
Non-recurrent & 17.80 & 18.89 & 22.25 & 23.85\\
\hline
Input-feeding & 19.93 & 21.41 & 25.83 & 27.18\\
\hline
\end{tabular}
\caption{\label{tbl:BLEUScores} Performance of our experimental system in BLEU on different standard WMT test sets.}
\end{table*}

\section{Empirical Analysis of Attention Behaviour}

We conduct our analysis using the two different attention models described in Section~\ref{sec:attentionModels}. Our first attention model is the global model without input-feeding as introduced by \newcite{DBLP_journals_corr_LuongPM15}. The second model is the input-feeding model \cite{DBLP_journals_corr_LuongPM15}, which uses recurrent attention.  Our NMT system is a unidirectional encoder-decoder system as described in \cite{DBLP_journals_corr_LuongPM15}, using 4 recurrent layers.
%=======
%\section{Empirical analysis of attention behaviour}
%\subsection{Setup}
%We conduct our analysis using the two different attention models described in section~\ref{sec:attentionModels}. Our first attention model is a similar implementation of the so called global model without input-feeding introduced in \cite{DBLP_journals_corr_LuongPM15}. The second model is an input-feeding model \cite{DBLP_journals_corr_LuongPM15}. These models has been separately used as the attention model in a 4-layer recurrent LSTM neural machine translation system. Our neural machine translation system is a unidirectional encoder-decoder system as described in \cite{DBLP_journals_corr_LuongPM15}. 

We trained the systems with dimension size of 1,000 and batch size of 80 for 20 epochs. The vocabulary for both source and target side is set to be the 30K most common words. The learning rate is set to be 1 and a maximum gradient norm of 5 has been used.  We also use a dropout rate of 0.3 to avoid overfitting.

%Our input-feeding model uses the same hidden dimension size and vocabulary for the both sides. We set the learning rate of the system to be 0.002, due to the fact that our input-feeding system requires much smaller learning rate to converge. Our input-feeding system converges much faster and to this end, we train it only for 2 epochs.  

\begin{table}[thb]
\centering
\small
\resizebox{\linewidth}{!}{
\begin{tabular}{|c|c|c|c|c|}
\hline
Data & \# of Sent & Min Len & Max Len & Average Len\\
\hline
WMT15 & 4,240,727 & 1 & 100 & 24.7\\
\hline
\end{tabular}}
\caption{\label{tbl:dataStats} Statistics for the parallel corpus used to train our models. The length statistics are based on the source side.}
\end{table}

\subsection{Impact of Attention Mechanism}
\label{sec:attnMechImpact}

We train both of the systems on the WMT15 German-to-English training data, see Table~\ref{tbl:dataStats} for some statistics. Table~\ref{tbl:BLEUScores} shows the BLEU scores \cite{papineni2002bleu} for both systems on different test sets.

Since we use POS tags and dependency roles in our analysis, both of which are based on words, we chose not to use BPE \cite{sennrichP16-1162} which operates at the sub-word level.
%Here, we chose not to use BPE \CM{REF?}, since as a result of using POS tags and dependency roles, which are word specific, in our experiments it introduces more complexity to the experiments. This is while it does not make a big difference on the focus of this study which is the difference of traditional alignments and attention. However, bpe could have been used by sharing the word specific features between all the splits of a word. }

%Note that we use the attention models of these two systems to provide an analysis of the attention behaviour and the reported translation performances are only to give a sense to the readers that these are valid translation systems, comparable to the state-of-the-art systems.

\begin{table}[thb]
\centering
\small
\resizebox{\linewidth}{!}{
\begin{tabular}{|c|c|c|c|}
\hline
 & non-recurrent & input-feeding & GIZA++ \\
 \hline
 AER & 0.60 & 0.37 & 0.31 \\
 \hline
\end{tabular}}
\caption{\label{tbl:alignmentErrorRate} Alignment error rate (AER) of the hard alignments produced from the output attentions of the systems with input-feeding and non-recurrent attention models. We use the most attended source word for each target word as the aligned word. The last column shows the AER for the alignment generated by GIZA++.}
\end{table}

\begin{table}[thb]
\centering
\small
\resizebox{\linewidth}{!}{
\begin{tabular}{|c|c|c|}
\hline
 & non-recurrent & input-feeding  \\
 \hline
 \makecell{Attention loss} & 0.46 & 0.25 \\
 \hline
\end{tabular}}
\caption{\label{tbl:averageAttnVsAlignLoss} Average loss between attention generated by input-feeding and non-recurrent systems and the manual alignment over RWTH German-English data.}
\end{table}

\begin{figure*}[thb]
\centering
\begin{subfigure}[t]{0.49\textwidth}
	\includegraphics[scale=0.12]{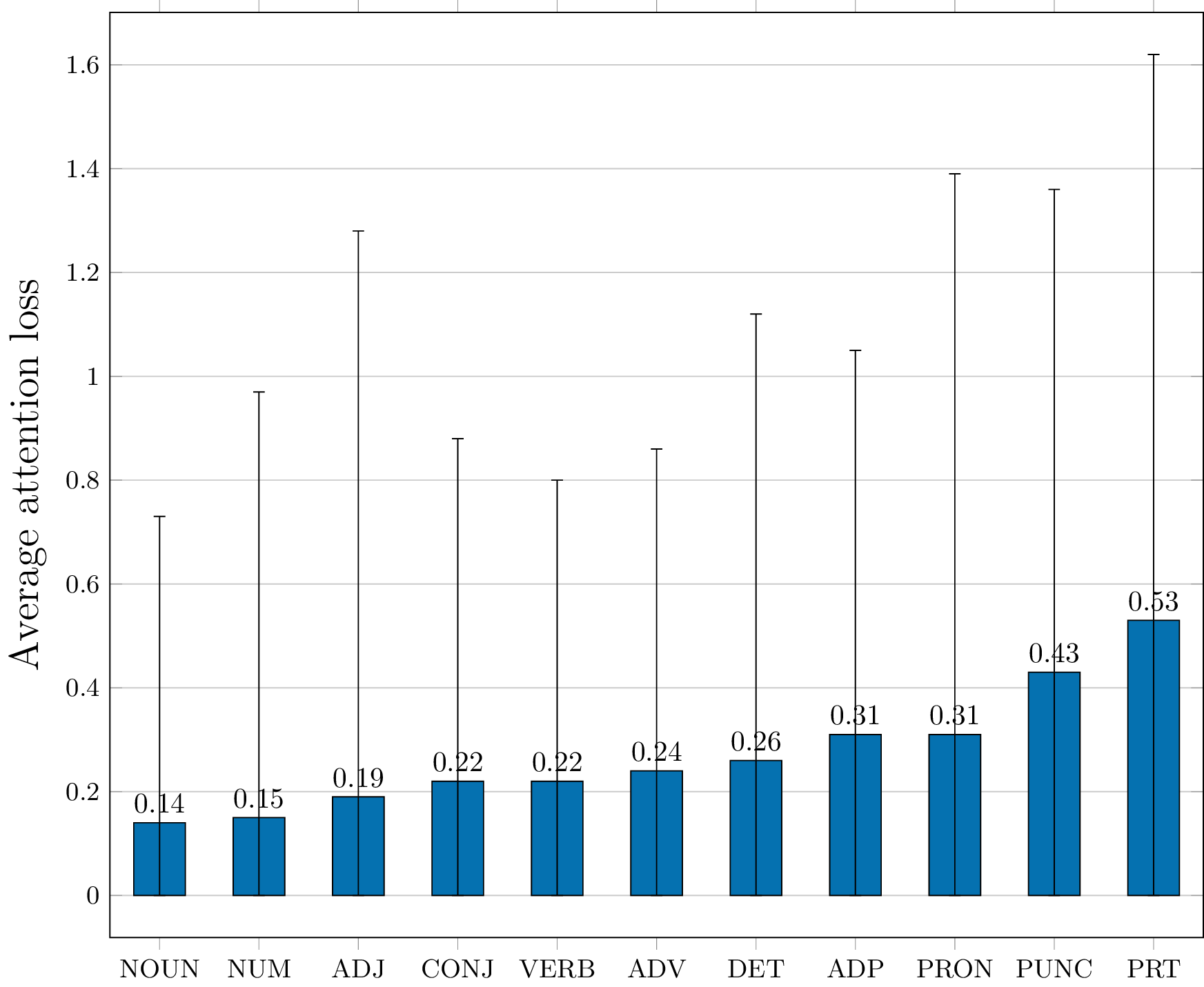}
	\caption{Average attention loss based on the POS tags of the target side.}
	\label{fig:attentionLoss}
\end{subfigure}%
~
\begin{subfigure}[t]{0.49\textwidth}
	\includegraphics[scale=0.12]{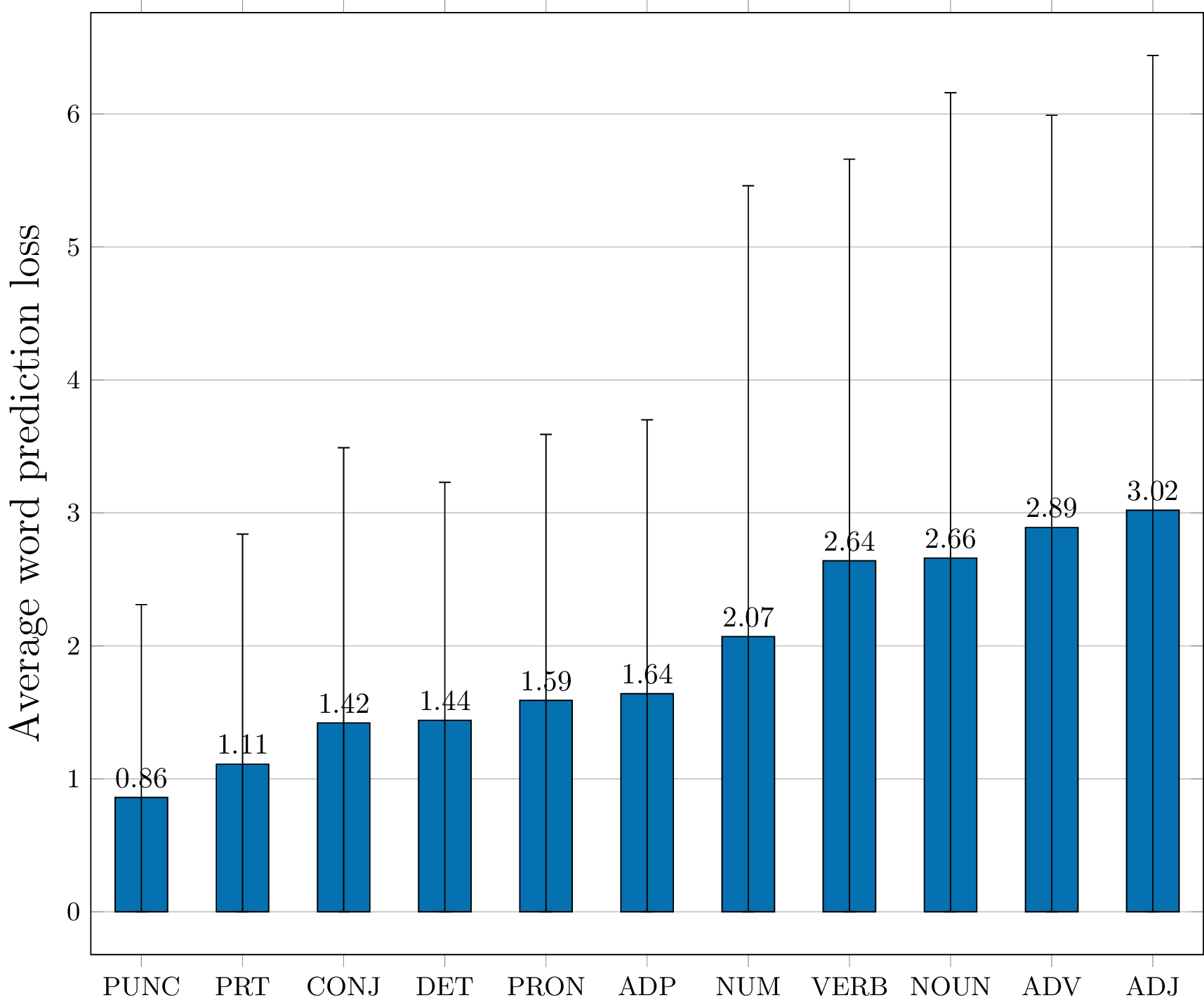}
	\caption{Average word prediction loss based on the POS tags of the target side.}
	\label{fig:transLoss}
\end{subfigure}
\caption{Average attention losses and word prediction losses from the input-feeding system.}
\end{figure*}

We report alignment error rate (AER) \cite{DBLP:conf/acl/OchN00}, which is commonly used to measure alignment quality, in Table~\ref{tbl:alignmentErrorRate} to show the difference between attentions and human alignments provided by RWTH German-English dataset. To compute AER over attentions, we follow \newcite{DBLP_journals_corr_LuongPM15} to produce hard alignments from attentions by choosing the most attended source word for each target word. We also use GIZA++ \cite{och2003systematic} to produce automatic alignments over the data set to allow for a comparison between automatically generated alignments and the attentions generated by our systems. GIZA++ is run in both directions and alignments are symmetrized using the grow-diag-final-and refined alignment heuristic.

%For the purpose of generating automatic alignment over a data set using GIZA++, we always concat the data set to the large bitext of 4M sentences over which we have trained our translation systems.

As shown in Table~\ref{tbl:alignmentErrorRate}, the input-feeding system not only achieves a higher BLEU score, but also uses attentions that are closer to the human alignments.

%has a closer attention to both gold alignments and the automatically generated alignments over RWTH data.

Table~\ref{tbl:averageAttnVsAlignLoss} compares input-feeding and non-recurrent attention in terms of attention loss computed using Equation~\ref{eq:attnLoss}. Here the losses between the attention produced by each system and the human alignments is reported. As expected, 
the difference in attention losses are in line with AER.

%Both comparisons in Table~\ref{tbl:averageAttnVsAlignLoss} and Table~\ref{tbl:alignmentErrorRate} consistently show that the input-feeding system generates attentions closer to the gold alignments and automatically generated alignments using tools like GIZA++, than what the non-recurrent attention model produces. 

The difference between these comparisons is that AER only takes the most attended word into account while attention loss considers the entire attention distribution. 

%At this point, one might be inclined to conclude that the closer the attention is to the word alignments the better the translation. In the following sections we investigate this further and will show that this is not always the case. 

\begin{table}[thb]
\centering
%\small
\resizebox{0.8\linewidth}{!}{
\begin{tabular}{|c|c|c|}
\hline
\multicolumn{1}{|c|}{Tag} & \multicolumn{1}{c|}{Meaning} & \multicolumn{1}{c|}{Example}\\
\hline
ADJ & Adjective & large, latest \\
ADP & Adposition & in, on, of \\
ADV & Adverb & only, whenever\\
CONJ & Conjunction & and, or\\
DET & Determiner & the, a\\
NOUN & Noun & market, system\\
NUM & Numeral & 2, two\\
PRT & Particle & 's, off, up\\
PRON & Pronoun & she, they\\
PUNC & Punctuation & ;, .\\
VERB & Verb & come, including\\
%X & Others & foreign words\\
\hline
\end{tabular}}
\caption{List of the universal POS tags used in our analysis.}
\label{tbl:POStags}
\end{table}  

%\subsection{Attention Patterns}
%
%Based on the results in Section~\ref{sec:attnMechImpact}, one might be inclined to conclude that the closer the attention is to the word alignments the better the translation. However \newcite{chen2016guided, liu-EtAl:2016:COLING, alkhouli-EtAl:2016:WMT} report mixed results by optimizing their NMT system with respect to word prediction and alignment quality. These findings warrant a more fine-grained analysis of attention. To this end, we chose to include POS tags in our analysis and study the patterns of attention based on POS tags of the target words. We chose POS tags because they exhibit some simple syntactic characteristics. We use the coarse grained universal POS tags \cite{petrov2012universal} given in Table~\ref{tbl:POStags}.

\subsection{Alignment Quality Impact on Translation}

Based on the results in Section~\ref{sec:attnMechImpact}, one might be inclined to conclude that the closer the attention is to the word alignments the better the translation. However, \newcite{chen2016guided, liu-EtAl:2016:COLING, alkhouli-EtAl:2016:WMT} report mixed results by optimizing their NMT system with respect to word prediction and alignment quality. These findings warrant a more fine-grained analysis of attention. To this end, we include POS tags in our analysis and study the patterns of attention based on POS tags of the target words. We choose POS tags because they exhibit some simple syntactic characteristics. We use the coarse grained universal POS tags \cite{petrov2012universal} given in Table~\ref{tbl:POStags}.

To better understand how attention accuracy affects translation quality, we analyse the relationship between attention loss and word prediction loss for individual part-of-speech classes. Figure~\ref{fig:attentionLoss} shows how attention loss differs when generating different POS tags. One can see that  attention loss varies substantially across different POS tags. In particular, we focus on the cases of NOUN and VERB which are the most frequent POS tags in the dataset. As shown, the attention of NOUN is the closest to alignments on average. But the average attention loss for VERB is almost two times larger than the loss for NOUN. 

Considering this difference and the observations in Section~\ref{sec:attnMechImpact}, a natural follow-up would be to focus on getting the attention of  verbs to be closer to alignments. However, Figure~\ref{fig:transLoss} shows that the average word prediction loss for verbs is actually smaller compared to the loss for nouns. In other words, although the attention for verbs is substantially more inconsistent with the word alignments than for nouns, the NMT system translates verbs more accurately than nouns on average.

\begin{figure}[thb]
\centering
\includegraphics[scale=0.12]{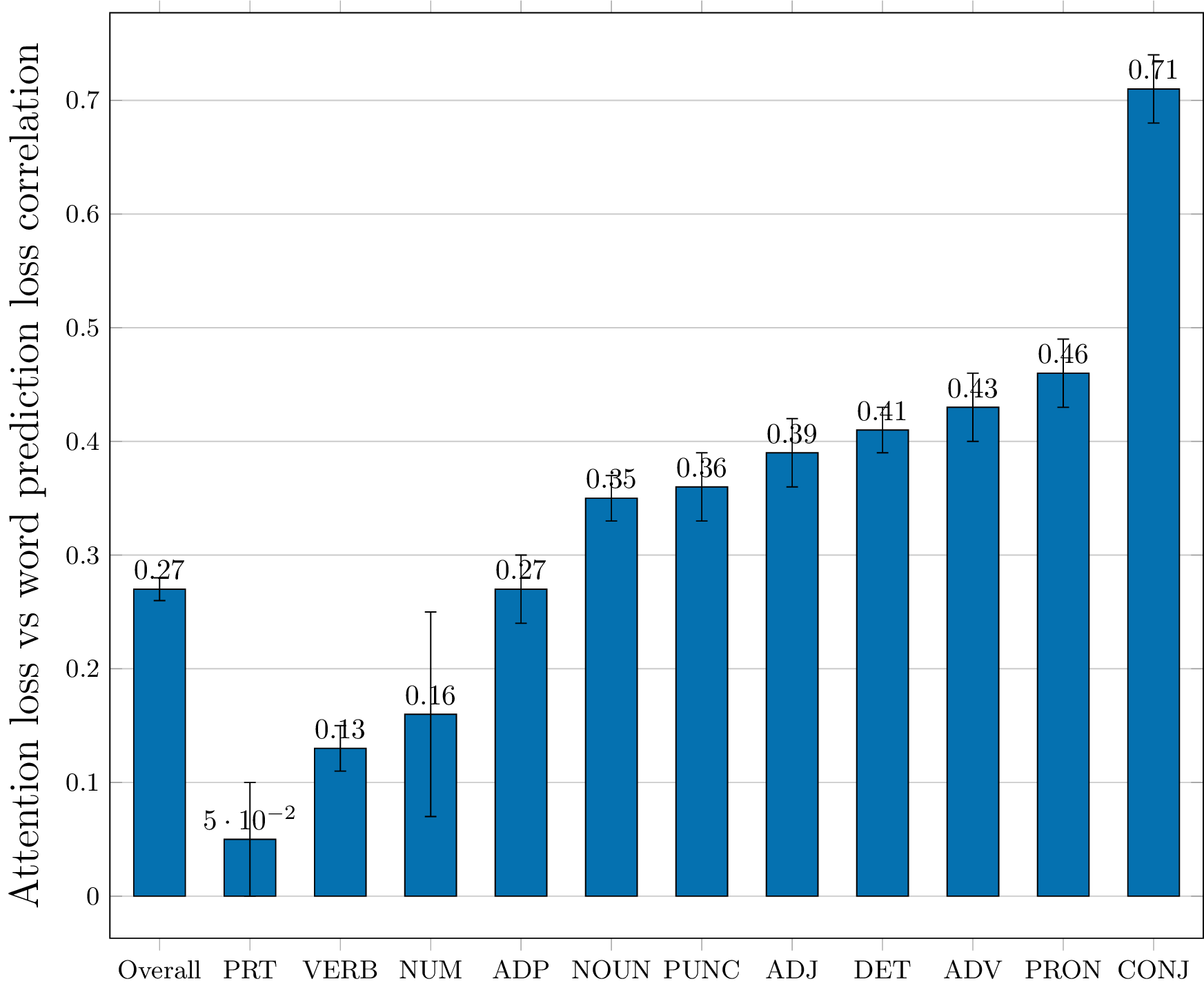}
\caption{Correlation between word prediction loss and attention loss for the input-feeding model.}
\label{fig:transLossVsAlignLoss}
\end{figure}

\begin{figure*}[thb]
\centering
\begin{subfigure}[t]{0.49\textwidth}
	\includegraphics[scale=0.12]{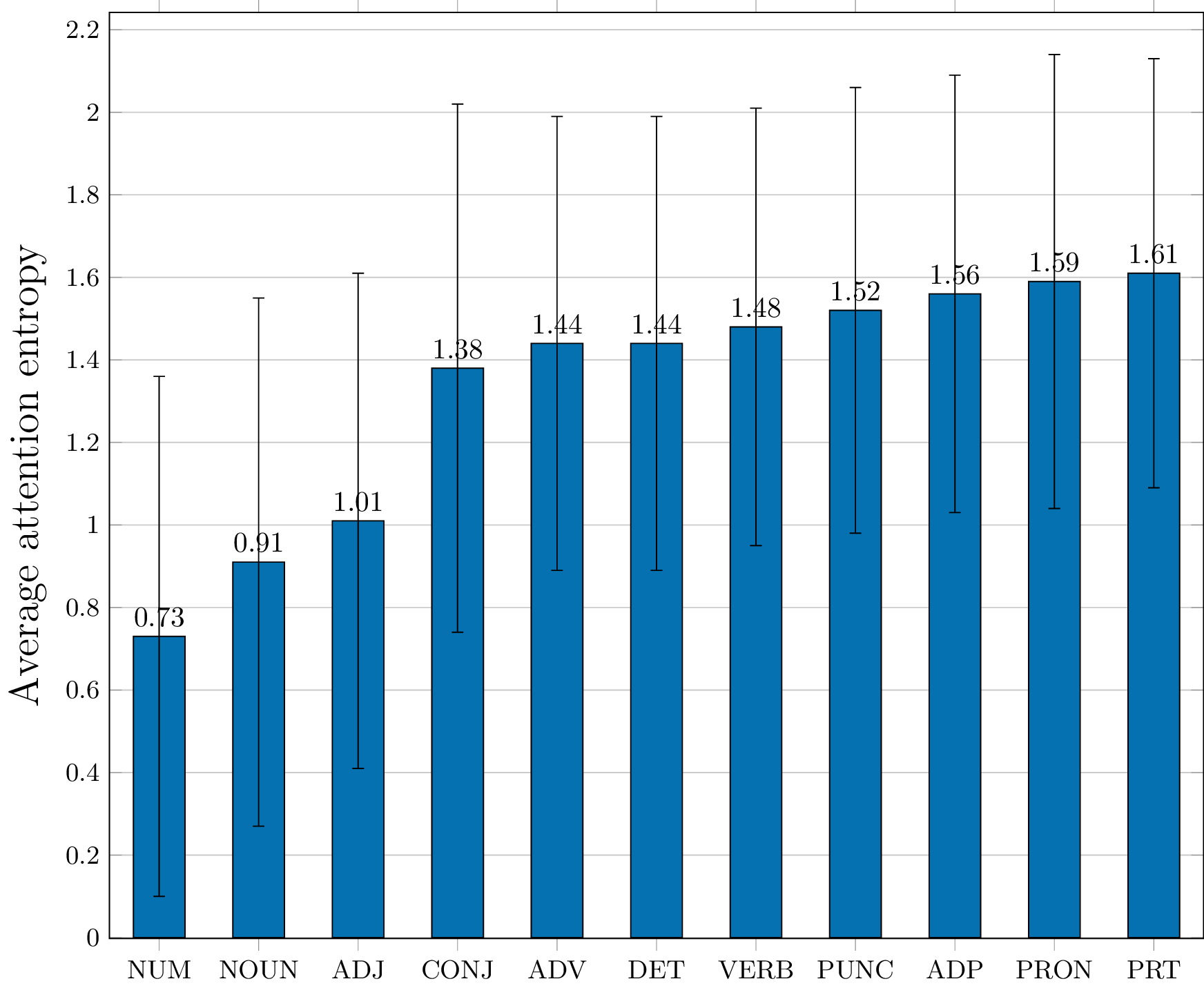}
	\caption{Average attention entropy based on the POS tags.}
	\label{fig:attnEntropy_a}
\end{subfigure}%
~
\begin{subfigure}[t]{0.49\textwidth}
	\includegraphics[scale=0.12]{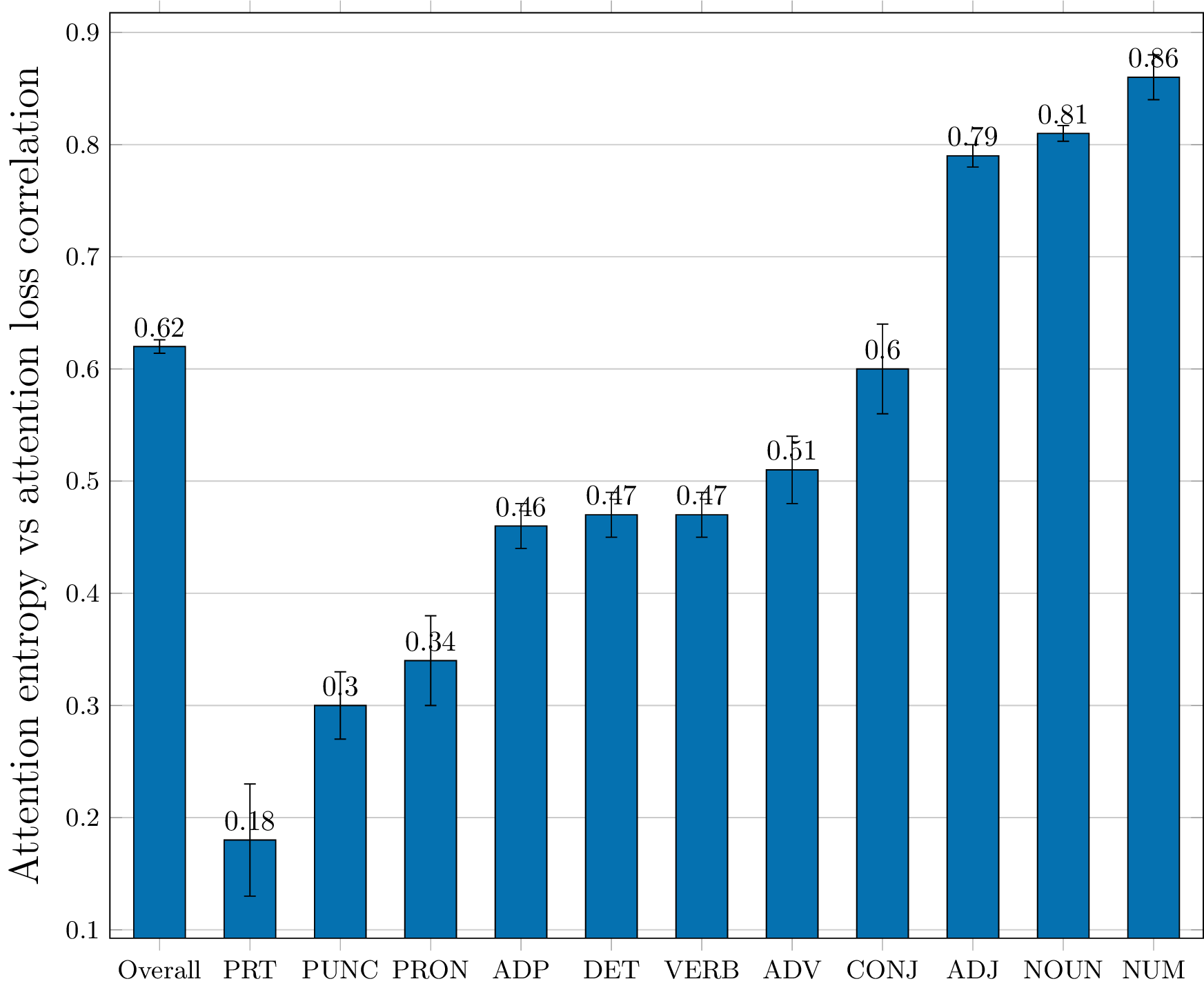}
	\caption{Correlation between attention entropy and attention loss.}
	\label{fig:attnEntropy_b}
\end{subfigure}
\caption{Attention entropy and its correlation with attention loss for the input-feeding system.}
\label{fig:AttnEntropyAndCorrelation}
\end{figure*}

To formalize this relationship we compute Spearman's rank correlation between word prediction loss and attention loss, based on the POS tags of the target side, for the input-feeding model, see Figure~\ref{fig:transLossVsAlignLoss}.

The low correlation for verbs confirms that attention to other parts of source sentence rather than the aligned word is necessary for translating verbs and that attention does not necessarily have to follow alignments. However, the higher correlation for nouns means that consistency of attention with alignments is more desirable. This could, in a way, explain the mixed result reported for training attention using alignments \cite{chen2016guided,liu-EtAl:2016:COLING, alkhouli-EtAl:2016:WMT}. Especially the results by \newcite{chen2016guided} in which large improvements are achieved for the e-commerce domain which contains many OOV product names and placeholders, but no or very weak improvements were achieved over common domains.  

%The first apparent result of the comparison given in the figure, is that the input-feeding system has generated less random points which has led to statistically significant increase in the correlation in almost all of the cases. In other words, we can easily see that the input-feeding system makes less error than the system without input-feeding. 

%The high correlation for CONJ in the case of input-feeding system shows that the consistency with the alignments is necessary for a better translation at least in case of translating into conjunctions. Note that the high correlation alone does not necessarily imply causal relationship between the two variables. But, taking into account the fact that the system translates attended words as well as the fact of the high correlation the necessity will follow.    

%A low correlation here does not necessarily mean that there is no relation between word prediction loss and attention loss in other cases, since it could be due to errors that the system still makes either in the attention or the translation under influence of other variables. However, it could also be evidence of that the system does not require its attention to exactly follow the alignments to be able to generate a good translation in those cases. In other words, it could mean that the attention is capturing other information rather than the alignments only, in those cases. 

\subsection{Attention Concentration}

In word alignment, most target words are aligned to one source word. The average number of source words aligned to nouns and verbs is 1.1 and 1.2 respectively. To investigate to what extent this also holds for attention we measure the attention concentration by computing the entropy of the attention distribution, see Equation~\ref{eq:attnEntropy}. 

Figure~\ref{fig:attnEntropy_a} shows the average entropy of attention based on POS tags. As shown, nouns have one of the lowest entropies meaning that on average the attention for nouns tends to be concentrated. This also explains the closeness of the attention to alignments for nouns. In addition, the correlation between attention entropy and attention loss in case of nouns is high as shown in Figure~\ref{fig:attnEntropy_b}. This means that attention entropy can be used as a measure of closeness of attention to alignment in the case of nouns.

\begin{figure}[thb]
\centering
\includegraphics[scale=0.12]{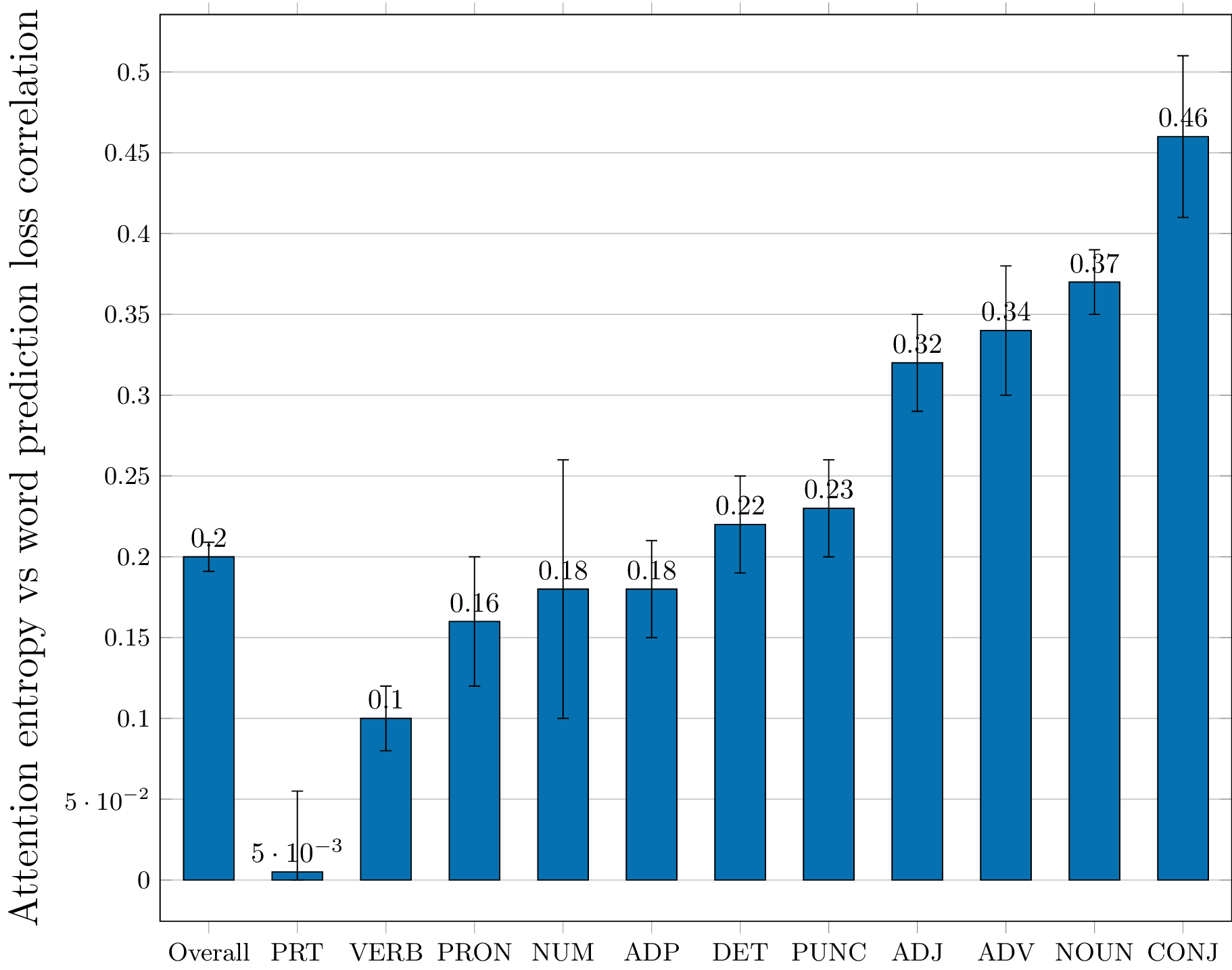}
\caption{Correlation of attention entropy and word prediction loss for the input-feeding system.}
\label{fig:transLossVsattnEntropy}
\end{figure}

The higher attention entropy for verbs, in Figure~\ref{fig:attnEntropy_a}, shows that the attention is more distributed compared to nouns. The low correlation between attention entropy and word prediction loss (see Figure~\ref{fig:transLossVsattnEntropy}) shows that attention concentration is not required when translating into verbs. This also confirms that the correct translation of verbs requires the systems to pay attention to different parts of the source sentence. 

\setcounter{footnote}{0}

\renewcommand{\arraystretch}{1.15}
\begin{table*}[thb]
\centering
\small
\begin{tabular}{|c|c|c|}
\hline
POS tag & roles(attention \%) & description\\
\hline
\multirow{4}{*}{NOUN} & punc(16\%) & Punctuations\tablefootnote{Punctuations have the role ``root" in the parse generated using ParZu. However, we use the pos tag to discriminate them from tokens having the role ``root".}\\
& pn(12\%) & Prepositional complements\\
& attr(10\%) & Attributive adjectives or numbers\\
& det(10\%) & Determiners\\
\hline
\multirow{5}{*}{VERB} & adv(16\%) & Adverbial functions including negation\\
& punc(14\%) & Punctuations\\
& aux(9\%) & Auxiliary verbs\\
& obj(9\%) & Objects\tablefootnote{Attention mass for all different objects are summed up.}\\
& subj(9\%) & Subjects\\
\hline
\multirow{3}{*}{CONJ} & punc(28\%) & Punctuations\\
& adv(11\%) & Adverbial functions including negation\\
& conj(10\%) & All members in a coordination\tablefootnote{Includes all different types of conjunctions and conjoined elements.}\\
\hline
\end{tabular}
\caption{The most attended dependency roles with their received attention percentage from the attention probability mass paid to the words other than the alignment points. Here, we focus on the POS tags discussed earlier.}
\label{tbl:dependencyRoles}
\end{table*}
\renewcommand{\arraystretch}{1}

Another interesting observation here is the low correlation for pronouns (PRON) and particles (PRT), see Figure~\ref{fig:transLossVsattnEntropy}. As can be seen in Figure~\ref{fig:attnEntropy_a}, these tags have more distributed attention comparing to nouns, for example. This could either mean that the attention model does not know where to focus or it deliberately pays attention to multiple, somehow relevant, places to be able to produce a better translation. The latter is supported by the relatively low word prediction losses, shown in the Figure~\ref{fig:transLoss}.

\subsection{Attention Distribution}

To further understand under which conditions attention is paid to words other than the aligned words, we study the distribution of attention over the source words. First, we measure how much attention is paid to the aligned words for each POS tag, on average. To this end, we compute the percentage of the probability mass that the attention model has assigned to aligned words for each POS tag, see Table~\ref{tbl:attnPercentage}.

\renewcommand{\arraystretch}{1.15}
\begin{table}[thb]
\centering
\small
\begin{tabular}{|c|c|c|}
\hline
POS tag & \makecell{attention to\\ alignment points \%} & \makecell{attention to\\ other words \%}\\
\hline
NUM & 73 & 27\\
NOUN & 68 & 32\\
ADJ & 66 & 34\\
PUNC & 55 & 45\\
ADV & 50 & 50\\
CONJ & 50 & 50\\
VERB & 49 & 51\\
ADP & 47 & 53\\
DET & 45 & 55\\
PRON & 45 & 55\\
PRT & 36 & 64\\
\hline
Overall & 54 & 46\\
\hline
\end{tabular}
\caption{Distribution of attention probability mass (in \%) over alignment points and the rest of the words for each POS tag.}
\label{tbl:attnPercentage}
\end{table}
\renewcommand{\arraystretch}{1}

One can notice that less than half of the attention is paid to alignment points for most of the POS tags. To examine how the rest of attention in each case has been distributed over the source sentence we measure the attention distribution over dependency roles in the source side. We first parse the source side of RWTH data using the ParZu parser \cite{sennrich-volk-schneider:2013:RANLP-2013}. Then we compute how the attention probability mass given to the words other than the alignment points, is distributed over dependency roles. Table~\ref{tbl:dependencyRoles} gives the most attended roles for each POS tag. Here, we focus on POS tags discussed earlier. One can see that the most attended roles when translating to nouns include adjectives and determiners and in the case of translating to verbs, it includes auxiliary verbs, adverbs (including negation), subjects, and objects. 

\section{Conclusion}

In this paper, we have studied attention in neural machine translation and provided an analysis of the relation between attention and word alignment. We have shown that attention agrees with traditional alignment to a certain extent. However, this differs substantially by attention mechanism and the type of the word being generated. We have shown that attention has different patterns based on the POS tag of the target word. The concentrated pattern of attention and the relatively high correlations for nouns show that training the attention with explicit alignment labels is useful for generating nouns. However, this is not the case for verbs, since the large portion of attention being paid to words other than alignment points, is already capturing other relevant information. Training attention with alignments in this case will force the attention model to forget these useful information. This explains the mixed results reported when guiding attention to comply with alignments \cite{chen2016guided, liu-EtAl:2016:COLING, alkhouli-EtAl:2016:WMT}.

\section*{Acknowledgments}

This research was funded in part by the Netherlands Organization for Scientific Research
(NWO) under project numbers 639.022.213 and 612.001.218.

\bibliography{./hghader}
\bibliographystyle{ijcnlp2017}

\end{document}